\documentclass[10pt,twocolumn,letterpaper]{article}

\usepackage{cvpr}              
\usepackage{amsmath}
\usepackage{amssymb}
\usepackage{mathtools}
\usepackage{bbding}
\usepackage{multirow}
\usepackage{makecell}
\usepackage{pifont}
\usepackage[table]{xcolor}
\usepackage{xcolor}
\usepackage{color}
\usepackage{colortbl}
\usepackage{tablefootnote}
\usepackage{pifont}
\usepackage{makecell}
\usepackage[T1]{fontenc}
\usepackage[most]{tcolorbox}
\newcommand{\ignore}[1]{}
\usepackage{stfloats}

\usepackage[most]{tcolorbox}
\tcbuselibrary{most}
\newcounter{promptid}[section]                        
\renewcommand{\thepromptid}{\thesection.\arabic{promptid}} 
\definecolor{darkorange}{RGB}{255, 140, 0}
\definecolor{lightgreen}{RGB}{145, 204, 117}
\definecolor{lightyellow}{RGB}{250, 200, 88}
\definecolor{lightred}{RGB}{238, 102, 102}
\definecolor{lightblue}{RGB}{115, 192, 222}
\newtcolorbox{promptbox}[2][Prompt]{
float,              
floatplacement=t!,
code={\refstepcounter{promptid}},     
title={Box~\thepromptid:~#1},
colback=black!5!white,
arc=5pt, 
boxrule=0.5pt,
fonttitle=\bfseries,
before upper={\footnotesize}, fontupper=\fontfamily{ptm}\selectfont,
colframe=#2, 
}

\newtcolorbox{promptboxbottom}[2][Prompt]{
    float,
    floatplacement=b!, 
    code={\refstepcounter{promptid}}, 
    title={Box~\thepromptid:~#1}, 
    colback=black!5!white,
    arc=5pt, 
    boxrule=0.5pt,
    fonttitle=\bfseries,
    before upper={\footnotesize}, 
    fontupper=\fontfamily{ptm}\selectfont,
    colframe=#2, 
}

\definecolor{cvprblue}{rgb}{0.21,0.49,0.74}
\usepackage[pagebackref,breaklinks,colorlinks,allcolors=cvprblue]{hyperref}

\def\paperID{}
\def\confName{CVPR}
\def\confYear{2026}

\title{Improving Vision-language Models with Perception-centric \\ Process Reward Models}

\author{
\textbf{Yingqian Min\textsuperscript{1,2\thanks{Equal contributions.}}}, \textbf{Kun Zhou\textsuperscript{3\footnotemark[1]}}, \textbf{Yifan Li\textsuperscript{1,2\footnotemark[1]}}, \textbf{Yuhuan Wu\textsuperscript{4}}, \textbf{Han Peng\textsuperscript{1}} \\ \textbf{Yifan Du\textsuperscript{1}}, \textbf{Wayne Xin Zhao\textsuperscript{1\thanks{Corresponding author.}}}, \textbf{Min Yang\textsuperscript{2}}, \textbf{Ji-Rong Wen\textsuperscript{1}} \\
 \textsuperscript{1}Gaoling School of Artificial Intelligence, Renmin University of China.\\
 \textsuperscript{2}Bytedance. 
 \textsuperscript{3}University of California, San Diego. \\
  \textsuperscript{4}The Hong Kong University of Science and Technology. \\
\texttt{yingqianm@ruc.edu.cn, batmanfly@gmail.com}
}

\begin{document}
\maketitle
\begin{abstract}
Recent advancements in reinforcement learning with verifiable rewards (RLVR) have significantly improved the complex reasoning ability of vision-language models (VLMs). However, its outcome-level supervision is too coarse to diagnose and correct errors within the reasoning chain. To this end, we propose Perceval, a process reward model (PRM) that enables token-level error grounding, which can extract image-related claims from the response and compare them one by one with the visual evidence in the image, ultimately returning claims that contain perceptual errors. Perceval is trained with perception-intensive supervised training data. We then integrate Perceval into the RL training process to train the policy models. Specifically, compared to traditional GRPO, which applies sequence-level advantages, we apply token-level advantages by targeting penalties on hallucinated spans identified by Perceval, thus enabling fine-grained supervision signals. In addition to augmenting the training process, Perceval can also assist VLMs during the inference stage. Using Perceval, we can truncate the erroneous portions of the model’s response, and then either have the model regenerate the response directly or induce the model to reflect on its previous output. This process can be repeated multiple times to achieve test-time scaling. Experiments show significant improvements on benchmarks from various domains across multiple reasoning VLMs trained with RL, highlighting the promise of perception-centric supervision as a general-purpose strategy. For test-time scaling, it also demonstrates consistent performance gains over other strategies, such as major voting. Our code and data will be publicly released at \url{https://github.com/RUCAIBox/Perceval}.

\end{abstract}
\section{Introduction}
\label{sec:intro}
Vision–language models (VLMs)~\cite{bai2023qwen, chen2024internvl, geminiteam2025geminifamilyhighlycapable} deliver strong results across tasks such as multimodal mathematics~\cite{wang2024measuring, lu2024mathvista}, chart analysis~\cite{masry-etal-2022-chartqa, liu2024ocrbench}, and general VQA~\cite{zhang2024mme}. However, they still falter on complex visual reasoning tasks, where multi-step chains of thought can be brittle and produce perceptual or logical mistakes~\cite{vstar, fu2024blink, mmstar}. To improve the performance, reinforcement learning with verifiable rewards (RLVR)~\cite{guo2025deepseek, team2025kimi, shao2024deepseekmathpushinglimitsmathematical} has become a widely used post-training strategy. Built on policy-gradient methods like PPO and GRPO, RLVR assigns outcome-level rewards to explicit reasoning traces  and optimizes the policy toward more consistent, robust multi-step visual reasoning.


Despite these advances, outcome-level supervision in RLVR is poorly matched to the inherently multi-step nature of visual reasoning. In fact, sequence-level rewards are too coarse to identify which perception or reasoning steps went wrong, creating a hard credit-assignment problem. In practice, VLMs often insert hallucinated objects or spatial relations and drift from the image context mid-chain~\cite{li2023evaluating, li2025analyzing, zhang2025modalities, agrawal2025towards, liu2025more}, but only the final reward offers little guidance about whether the failure arose from visual grounding or subsequent logic. Thus, the sparse-reward regime ultimately bottlenecks RLVR's gains on VLMs~\cite{zhang2025vl}.


To overcome the sparse-reward limitation, we introduce a process reward model (PRM) that supervises intermediate steps rather than only the final outcome~\cite{wang2024math}. Prior work shows that PRMs can effectively guide both training and inference by rewarding stepwise, chain-of-thought correctness~\cite{lightman2023let, zheng2025survey}.
However, building a high-quality PRM is difficult because step-level annotations are expensive and some steps are only verifiable after later derivations, complicating labeling and consistency~\cite{zhang2025lessonsdevelopingprocessreward, khalaf2025inferencetimerewardhackinglarge}.
Fortunately, in visual reasoning many intermediate steps are perceptual claims (\eg, objects, attributes, or spatial relations) that can be grounded directly in the image, enabling automatic checks for ``image–text misalignment'' (hallucination). 
Therefore, it is promising to develop a perception-centric PRM that detects and explains such misalignments to provide fine-grained feedback, alleviating sparse-reward issue and improving learning of the reasoning ability.

To operationalize this, we first define a perception-level error-finding schema for a perception-centric PRM. We curate training queries from perception-intensive settings—such as goal-directed visual search and referring-expression grounding—and use a strong LLM to produce structured annotations that mark image–text misalignments (hallucinatory spans and their visual counter-evidence). After supervised fine-tuning on this corpus, the PRM can reliably flag hallucinations that arise within multi-step rationales and return well-structured feedback. 
Building on this, we integrate the PRM into RLVR by decomposing the sequence-level advantage and assigning fine-grained, token-level penalties to spans identified as hallucinatory, yielding more precise credit assignment than GRPO alone. 
Finally, based on PRM’s structured outputs, we employ a simple Truncation–Regeneration loop at inference. In this way, suspect spans are pruned and regenerated, trading a bit more compute for stronger factual grounding. 

Experimental results demonstrate that, compared to direct GRPO, our training method significantly enhances the model's perceptual capabilities, boosting performance on perception-centric tasks. Furthermore, we observe a surprising and significant generalization effect: even without applying PRM supervision during the training for complex reasoning tasks, this foundational improvement in perception nonetheless generalizes, leading to a comprehensive enhancement of the model's overall reasoning abilities.

Our main contributions are as follows:
\begin{itemize}
\item We propose a novel, perception-centric process reward model (PRM) that can explicitly identify perception errors in the reasoning process.

\item We introduce a fine-grained, token-level advantage re-allocation framework that integrates our PRM with GRPO, to solve the sparse reward issue.


\item We design a test-time iterative refinement strategy that leverages our PRM to actively detect and correct perceptual errors from the policy model.
\end{itemize}

\section{Preliminary}
\label{sec:preliminary}
We introduce foundational concepts and notations used throughout this paper: the architecture of vision-language models (VLMs), the reinforcement-learning framework with verifiable rewards (RLVR) which our method builds on, and our problem statement for designing a perception-centric process reward model.

\paragraph{Vision-Language Models.}
A vision-language model (VLM) accepts multimodal input, typically an image $v$ and a text query $q$, and generates the text output $o$, denoted as $\pi_{\theta}(o|q,v)$. 
For reasoning tasks, the text output is generally a chain of language reasoning steps.
Typical architecture combines a visual encoder (\eg ViT) to embed $I$ and a large language model~(LLM) to decode the output. Typically, the two modalities are linked via a connection layer.

\paragraph{Reinforcement Learning with Verifiable Rewards.}
RL with verifiable rewards (RLVR) has become the key technique to improve the performance of VLMs in reasoning tasks~\cite{yu2025perceptionr1pioneeringperceptionpolicy}.
It aims to train the VLM to not only generate plausible outputs but also satisfy measurable criteria (\eg, correctness, spatial consistency).
One algorithm is Group Relative Policy Optimization (GRPO)~\cite{shao2024deepseekmathpushinglimitsmathematical}: given the input prompt $q$ and image $v$, a reference policy $\pi_{\theta}(o|q,v)$ samples multiple responses $\{o_i\}$. Each response will be assigned with a scalar reward $R_i$ from the verified function or reward model. The advantage of the $i$-th response is calculated by normalizing its reward relative to the group:
\begin{equation}
\label{eq:grpo_advantage}
\hat{A}_{i} = \frac{R_i - \text{mean}(\{R_j\}_{j=1}^G)}{\text{std}(\{R_j\}_{j=1}^G)}
\end{equation}
Note that this advantage $\hat{A}_i$ is a \textbf{sequence-level} signal, which is constant for all tokens within the $i$-th response.
Hence, GRPO optimizes a clipped surrogate objective to update the policy $\pi_\theta$ based on the advantage:
\begin{equation}
\label{eq:grpo_objective}
\begin{multlined}
\small
J(\theta) = \mathbb{E}_{(q, \{o_i\}) \sim \pi_{\theta}} \Biggl[
\frac{1}{G} \sum_{i=1}^{G} \sum_{t=1}^{|o_i|} \min \Big( r_{i,t}(\theta)\hat{A}_{i}, \\
\text{clip}(r_{i,t}(\theta), 1-\epsilon, 1+\epsilon)\hat{A}'_{i,t} \Big)
- \beta D_{KL}(\pi_\theta || \pi_{ref}) \Biggr]
\end{multlined}
\end{equation}
where $\epsilon$ is the clipping hyperparameter and $r_{i,t}(\theta)$ is the importance sampling ratio for token $t$.

\paragraph{Problem Statement.}
A key limitation of reinforcement learning with verifiable rewards (RLVR) is \emph{reward sparsity}: conventional approaches provide a single scalar reward only at the end of the reasoning chain, so each token or step is credited equally regardless of its individual correctness or contribution. This coarse, sequence-level feedback makes it difficult to correct localized errors in perception or reasoning and undermines the model’s ability to generalize robustly.
To overcome this, we propose training a \emph{perception-centric process reward model (PRM)} that evaluates intermediate perceptual outputs and produces step-wise feedback. 
Concretly, the PRM checks whether the model’s perception content in response~(\eg, a grounding, visual feature, or intermediate state) is correct relative to the input $v,q$, and generate structured outputs that can be used to provide fine-grained supervision.
During inference, the PRM can be used to guide the selection of intermediate steps.
During training, by designing proper learning objective with the PRM, we encourage correct intermediate perceptual reasoning, enabling more fine-grained supervision for effective learning.

\section{Methodology}
\label{sec:method}




\begin{figure*}[t]
    \centering
    \vspace{-10pt}
    \includegraphics[width=\textwidth]{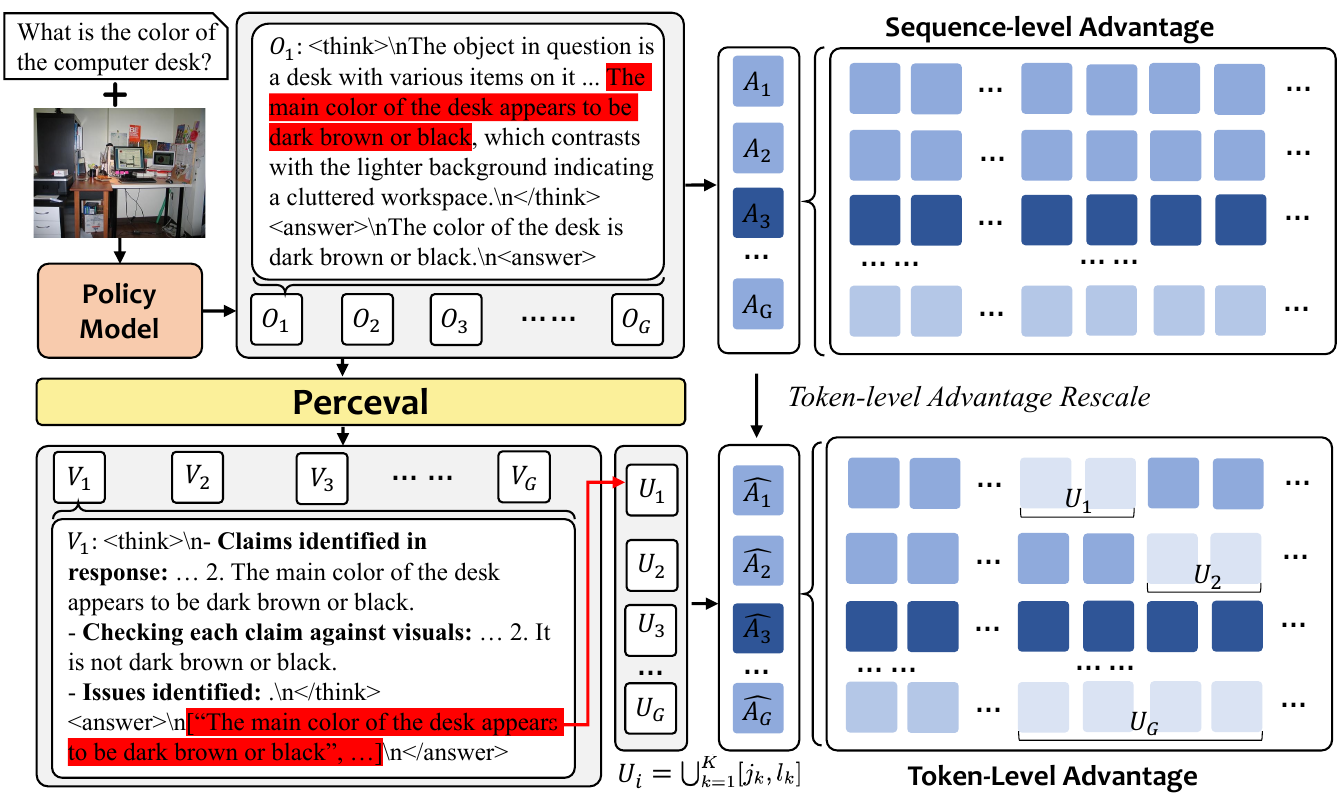}
    \caption{An overview of our Process-Supervised GRPO framework. For each generated response, we use the Perceval to create a token-level penalty mask. This mask is used to calculate a fine-grained token-level advantage, which is then incorporated into the GRPO objective to penalize hallucinatory tokens and improve the model's perceptual grounding.}
    \label{fig:ps-grpo-overview}
\end{figure*}

In this section, we devise our perception-centric process reward model for providing fine-grained, process-level supervision to guide VLMs.
We first introduce the design and how to train the PRM, and then present how to integrate it with RLVR during training and how to perform test-time scaling with PRM guidance.

\subsection{Perception-Centric Process Reward Model}
\label{subsec:pc-prm}

To overcome the sparse supervision issue, we propose \textsc{Perceval} (\textbf{Perc}eption-centric process reward \textbf{eval}uation model), which serves as an external, fine-grained, and interpretable critic for guiding VLM policy.

\paragraph{Error-finding Schema Design.}
Given a tuple of image, text query, and model's response $\langle v,q,o\rangle$, \textsc{Perceval} generates a structured verification $V$ to assess the factual consistency 
with respect to $v$ (conditioned on $q$). To improve reliability, \textsc{Perceval} follows the well-known \emph{think-then-answer} paradigm~\cite{guo2025deepseek}: it first analyzes each claim and outputs the thought process within \texttt{<think>...</think>}, where each statement in $o$ is evaluated for consistency with the visual evidence in $v$. Based on these analyses, \textsc{Perceval} provides the final decision wrapped in \texttt{<answer>...</answer>}. If no perceptual errors are found, the final answer is simply \texttt{"The response is correct."}; otherwise, the answer is formatted as a Python list containing the exact strings from $o$ that are identified as errors.

\paragraph{Process Reward Model Training.}
We train \textsc{Perceval} using a dataset constructed via a four-stage pipeline:

$\bullet$ \emph{Query selection}: to emphasize perceptual grounding, we primarily source the images and queries from visual search datasets \cite{vstar, deepeyes} that require locating specific objects in large images, and we include a small proportion from other domains (\eg, mathematical reasoning and general understanding~\cite{sophiavlr1}) to preserve breadth;

$\bullet$ \emph{Rollout generation}: based on the images and queries, we use an open-source VLM (\eg, Qwen2.5-VL-7B) to produce responses, whose imperfect perceptual alignment yields realistic hallucinations as negative examples;

$\bullet$ \emph{Automated annotation and verification}: for each response, we adopt a strong models (\eg, Gemini-2.5-Pro) to perform hallucination-focused, step-by-step checks. The generated annotations follow our designed format.

$\bullet$ \emph{Supervised fine-tuning}: we fine-tune the \textsc{Perceval} backbone with a standard SFT objective on the aggregated data to emulate detailed, perception-centric verification and produce the prescribed structured output.

\subsection{RLVR with Process-level Supervision}
\label{subsec:integration}
Building on \textsc{Perceval}, we revise the GRPO objective to support \emph{process-level} supervision by replacing the coarse sequence-level advantage $\hat{A}_i$ (Eq.~\ref{eq:grpo_advantage}) with a \emph{token-level} advantage $\hat{A}'_{i,t}$. The key change is to let advantage computation accept per-token signals so that perceptual errors within a response are directly penalized during learning. 
To achieve it, for each response, we use \textsc{Perceval} to identify the token spans that realize perception-induced hallucinations, and then re-assign advantages for those tokens to provide a reduced (or more negative) learning signal. 

Given a response $o_i$ of length $L_i$ and the \textsc{Perceval} verification, we parse the \texttt{<answer>} content and select the identified problematic substrings.
We locate each substring in $o_i$ via exact string match to obtain its token span $[j_k,l_k]$, and define $U_i=\bigcup_{k=1}^K [j_k,l_k]$. From $U_i$ we construct a binary mask $M_i=[m_{i,1},\dots,m_{i,L_i}]$ with $m_{i,t}=1$ if $t\in U_i$ and $0$ otherwise. Then, we modulate the sequence-level signal with this mask to form the token-level advantage:
\begin{equation}
\label{eq:token_advantage}
\hat{A}'_{i,t} \coloneqq \hat{A}_i - \alpha \cdot m_{i,t} \cdot \lvert \hat{A}_i \rvert,
\end{equation}
where $\alpha\in[0,1]$ controls penalty strength. Thus, correct tokens ($m_{i,t}=0$) keep $\hat{A}'_{i,t}=\hat{A}_i$, while hallucination tokens ($m_{i,t}=1$) are downweighted: when $\hat{A}_i>0$, $\hat{A}'_{i,t}=\hat{A}_i(1-\alpha)$; when $\hat{A}_i<0$, $\hat{A}'_{i,t}=\hat{A}_i(1+\alpha)$, making the penalty stronger. Finally, we substitute $\hat{A}'_{i,t}$ into the GRPO objective in Eq.~\ref{eq:grpo_objective} to add the process supervision.
Such a way injects direct, token-level corrective pressure into GRPO, which preserve sequence-level preferences while explicitly suppressing ungrounded content.

\subsection{Test-Time Scaling with PRM Guidance}
\label{subsec:test_time_scaling}
Beyond training-time use, \textsc{Perceval} (our perception-centric PRM) enables test-time scaling by supplying targeted error-correction during inference. We introduce two pragmatic refinement loops:

\paragraph{Truncate–then–Regenerate.}
When \textsc{Perceval} detects an erroneous claim, it returns the offending span in the model’s rationale. We truncate the hypothesis \emph{before} the first token of that span, preserving only the verified prefix as context. The policy model then continues to \emph{regenerate} the answer following this cleaned prefix. As the original image and question are given, the VLM just needs to resample the detected hallucinated part, without rewriting verified content. This truncate–continue cycle repeats until no new errors are flagged or a maximum of $k$ iterations is reached. The iteration cap $k$ bounds latency while typically yielding large accuracy gains with only a few refinement steps.

\paragraph{Truncate–Thinking–then–Regenerate.}
To further encourage self-correction, we augment the above method with a lightweight guidance for thinking. After truncating at the error, we append a brief thinking prompt in \textsc{Perceval}’s output, \eg, ``\texttt{Wait, I need to reconsider this reasoning more carefully: the mug is \emph{not} on the brick in the image.}'', which guides the model to think and then regenerate from the augmented context. The added thinking process enables self-reflection on the failure mode (object/attribute/spatial mismatch), improving the likelihood that the continuation repairs the specific misalignment. As with Truncate–then–Regenerate, we iterate up to $k$ times or stop early when no further errors are found, trading modest extra compute for stronger factual grounding.
\section{Experiment}
\label{sec:experiment}

\subsection{Experimental Setup}

\paragraph{Benchmarks.}
We select multiple visual reasoning benchmarks, covering visual search, perception-intensive reasoning, mathematical and chart-based reasoning.
\begin{enumerate}
    \item \textbf{V*~(V-Star)}~\cite{vstar}: introduces an LLM-guided visual search mechanism and a dedicated benchmark, to assess models’ ability to localize and reason about small, target objects within information-dense images. It contains 191 high-resolution images with two subtasks, \ie attribute recognition and spatial-relation reasoning that require precise grounding before reasoning.
    \item \textbf{MME-RealWorld}~\cite{zhang2024mme}: targets practical applications across five domains (OCR-in-the-wild, remote sensing, diagrams/tables, monitoring, autonomous driving). We use its subset MME-RealWorld-Lite for testing.
    \item \textbf{BLINK}~\cite{fu2024blink}: reframes 14 classic computer-vision tasks (\eg relative depth, visual correspondence, image forensics, multi-view reasoning) into 3,807 multiple-choice items to probe foundational perceptual skills that resist purely linguistic mediation.
    \item \textbf{MMStar}~\cite{mmstar}: compiles 1,500 carefully selected, human-curated samples to probe six core capability areas along 18 fine-grained axes, focusing on cases where vision is indispensable (rather than solvable by text priors).
    \item \textbf{RealWorldQA}~\cite{xai2024grok}: contains 700 images captured from vehicles and other real-world settings, each paired with a question and an easily verifiable answer. 
    \item \textbf{MathVista}~\cite{lu2024mathvista}: aggregates 6,141 examples from 28 existing multimodal sources and three new sets (IQTest, FunctionQA, PaperQA) to test numeracy, geometry/diagram understanding, tables/plots, and compositional visual-math reasoning.
    \item \textbf{MATH-Vision}~\cite{wang2024measuring}: offers 3,040 problems sourced from real competitions, spanning 16 mathematical disciplines and five difficulty levels, each embedded in a visual context (figures, diagrams, plots).
    \item \textbf{ChartQA}\cite{masry-etal-2022-chartqa}: contains 9.6K human-written and 23.1K generated questions over diverse chart types, requiring both visual parsing and table/logic operations.
\end{enumerate}

\paragraph{Baselines.}
We compare our methods with multiple reasoning-oriented VLMs. 
\begin{enumerate}
    \item \textbf{VLM-R1}~\cite{shen2025vlm}: extends R1-style RLVR to VLMs by leveraging tasks with deterministic visual ground truth.
    \item \textbf{LMM-R1}~\cite{peng2025lmm}:leverages text-only data with rule-based RL and multimodal generalization training to transfer gains to vision reasoning task.
    \item \textbf{R1-VL}~\cite{zhang2025r1vllearningreasonmultimodal}: proposes StepGRPO, replacing sequence-level rewards with dense step-wise rule-based rewards to stabilize visual reasoning ability learning.
    \item \textbf{Perception-R1}~\cite{yu2025perceptionr1pioneeringperceptionpolicy}: targets perception-heavy tasks and utilizes GRPO with perception-oriented rewards.
    \item \textbf{Jigsaw-R1}~\cite{wang2025jigsawr1studyrulebasedvisual}: is first trained on jigsaw puzzles data to improve generation, and then visual reasoning datasets.
    \item \textbf{DeepEyes}~\cite{deepeyes}: is end-to-end trained with RL to think with images and interleaves the visual grounding step inside the whole reasoning process.
    \item \textbf{PixelReasoner}~\cite{wang2025pixelreasonerincentivizingpixelspace}: adopts pixel-space reasoning (\eg zoom and crop) and a two-phase training: fine-tuning on synthesized data, then curiosity-driven RL.
    \item \textbf{Vision-R1}~\cite{huang2025vision}: cold-starts via a synthetic dataset, then applies GRPO with a hard formatting reward and a progressive thinking suppression training strategy.
    \item \textbf{VL-Rethinker}~\cite{wang2025vl}: uses GRPO with selective sample replay to mitigate vanishing advantages and adds forced rethinking triggers to elicit reflection.
    \item \textbf{VLAA-Thinker}~\cite{chen2025sftrlearlyinvestigation}: is trained using mixed verifiable rewards on multimodal CoT dataset with GRPO.
    \item \textbf{OpenVLThinker}~\cite{deng2025openvlthinkercomplexvisionlanguagereasoning}: iterate fine-tuning on distilled data and RL for improvement until convergence.
    \item \textbf{MM-Eureka}~\cite{meng2025mmeurekaexploringfrontiersmultimodal}: scales up the training data for rule-based RL in multimodal settings. 
\end{enumerate}
\paragraph{Implementation Details.}
\label{sec:implementation_details}
We select Qwen2.5-VL as the backbone for both reward and policy models. We first train two versions of \textsc{Perceval} of 3B and 7B sizes, following the procedures outlined in section~\ref{subsec:pc-prm}, and then correspondingly train two policy models of the same sizes using the proposed method. As for the training data, the supervised fine-tuning data are collected from DeepEyes~\cite{deepeyes} and SophiaVL-R1~\cite{sophiavlr1}, each of which is rolled out 3 times using the backbone models. The RL training data is also derived from \cite{deepeyes}, with the primary objective of enhancing the model's perception capabilities, while also containing a subset of general-purpose reasoning data. Consequently, during the RL training phase, we implement a conditional strategy: \textsc{Perceval} is used only on perception-related data to perform fine-grained advantage rescaling. For all other training data (\eg, mathematical reasoning), no additional intervention is applied, and we revert to using direct GRPO. This experimental design allows us to investigate whether fine-grained supervision focused on perception tasks can generalize and yield performance gains in other domains.

\paragraph{Evaluation Setup.}
To ensure fair and reproducible evaluation, we establish a unified evaluation pipeline. We employ greedy decoding for all models and utilize the same prompt template to collect responses. We then extract the final answer following the official procedures of each benchmark. Finally, the accuracy is determined through a two-stage judging process: we first apply an exact match (EM) judge for each extracted answer against the ground truth. For any answer that does not match, a robust judge model (\ie GPT-4o-mini) is utilized for a final verification to account for minor formatting variations. Additionally, we report the relaxed accuracy for ChartQA \cite{masry-etal-2022-chartqa}, aligned with the official evaluation of the benchmark, which uses the methodology of PlotQA \cite{Methani_2020_WACV}.

\begin{table*}[tbp]
    \small
    \centering
    \caption{Main results on multimodal benchmarks regarding visual search, perception-intensive reasoning and math\&chart tasks. MRW and RWQA denote MME-RealWorld and RealWorldQA, respectively. Best and second best results in each group are highlighted in bold and underlined, respectively. $^*$ indicates models capable of calling tools.}\label{tab:main}
    \scalebox{0.9}{
    \begin{tabular}{lccccccccccc}
    \toprule
    \multirow{2.5}{*}{\textbf{Models}} & \multirow{2.5}{*}{\textbf{\#Param}} &\multicolumn{3}{c}{\textbf{Visual Search}} & \multicolumn{4}{c}{\textbf{Perception-intensive Reasoning}} & \multicolumn{3}{c}{\textbf{Math \& Chart }} \\
    \cmidrule(lr){3-5}
    \cmidrule(lr){6-9}
    \cmidrule(lr){10-12}
    & &\textbf{V$^*_{\text{attr}}$} & \textbf{V$^*_{\text{pos}}$} & \textbf{V$^*_{\text{all}}$} & \textbf{BLINK} & \textbf{MMStar} & \textbf{MRW} & \textbf{RWQA} & \textbf{MathVision} & \textbf{MathVista} & \textbf{ChartQA}  \\
    \midrule
    VLM-R1~\cite{shen2025vlm} &3B &  75.65 & 67.11 & 72.25 & 46.25 & \textbf{56.7} & 42.3 & 61.5 & 21.71 & \underline{65.1} & 83.48  \\
    LMM-R1~\cite{peng2025lmmr1empowering3blmms} &3B  &  46.09 & 53.95 & 49.21 & 46.60 & \textbf{56.7} & 35.8 & 58.7 & \underline{24.47} & 63.5 & \underline{85.04} \\
    R1-VL~\cite{zhang2025r1vllearningreasonmultimodal} &2B  &  59.13 & 57.89 & 58.64 & 42.81 & 38.9 & 32.4 & 49.2 & 10.20 & 47.0 & 68.08 \\
    Perception-R1~\cite{yu2025perceptionr1pioneeringperceptionpolicy} & 3B &  57.39 & 48.68 & 53.92 & 46.44 & 54.8 & 37.5 & 55.8 & 22.03 & 58.1 & 81.60 \\
    Jigsaw-R1~\cite{wang2025jigsawr1studyrulebasedvisual} &3B  &  72.17 & 65.79 & 69.63 & 45.01 & 54.4 & 42.2 & 57.9 & 19.40 & 61.0 & 84.60 \\
    \midrule
    Qwen2.5-VL~\cite{Qwen2.5-VL} &3B  &  57.39 & 65.79 & 60.73 & 46.94 & 52.1 & 41.7 & \underline{63.4} & 21.40 & 61.6 & 83.12 \\
    + GRPO &3B  &  \underline{86.95}&  \underline{69.73}&  \underline{80.10}&  \textbf{49.13}&  55.3 &  \underline{46.8} & 62.1 & 23.36 & \underline{65.1} & 83.32  \\
    \rowcolor{gray!20}
    + Ours &3B  &  \textbf{90.43} &  \textbf{72.37} &  \textbf{83.25} &  \underline{48.75} &  \underline{55.8} & \textbf{47.6}  & \textbf{64.9}  & \textbf{26.32} & \textbf{65.6} & \textbf{86.48} \\
    \midrule
    \midrule
    DeepEyes~\cite{deepeyes}$*$ &7B &  \textbf{91.30} & 81.58  & \textbf{87.43} & 50.98 & 62.7 & 46.5 & \underline{67.0} & 12.50 & 69.9 & 75.84  \\
    Pixel-Reasoner~\cite{peng2025lmmr1empowering3blmms}$*$ &7B  & -  & - & $84.30$ & 51.10 & 63.1 & 43.5 & 64.0 & 22.03 & 69.4 & 77.36  \\
    Vision-R1~\cite{huang2025vision} &7B  &  - & - & -& 49.72 & 55.4 & \underline{49.6} & 64.1 & \textbf{36.18} & 71.3 & 83.36 \\
    VL-Rethinker~\cite{wang2025vl} &7B  &  54.78 & 59.21 & 56.54 & 49.91 & \textbf{64.0} & 38.8 & 64.0 & 31.91 & \textbf{72.6} & 85.60 \\
    VLAA-Thinker~\cite{chen2025sftrlearlyinvestigation} &7B  &  43.47 & 52.63 & 47.12 & 49.38 & \textbf{64.0} & 48.0 & 62.0 & 27.96 & 70.3 & \underline{85.36} \\
    R1-VL~\cite{zhang2025r1vllearningreasonmultimodal} &7B  &  47.83 & 67.11 & 55.50 & 47.19 & 55.5 & 40.5 & 59.5 & 22.37 & 64.1 & 82.80 \\
    OpenVLThinker~\cite{deng2025openvlthinkercomplexvisionlanguagereasoning} &7B  &  76.52 & 80.26 & 78.01 & 51.36 & 62.8 & 59.1 & 66.5 & 32.57 & 71.1 & \textbf{89.00} \\
    MM-Eureka~\cite{meng2025mmeurekaexploringfrontiersmultimodal} &7B  &  42.61 & 56.58 & 48.17 & 50.23 & 63.4 & 46.4 & 62.3 & 32.23 & \underline{72.4} & 82.36 \\
    \midrule
    Qwen2.5-VL~\cite{Qwen2.5-VL} &7B  &  60.87 & 64.47 & 62.30 & 48.56 &	62.3 &	43.0 &	60.6 &	26.97	&70.2 &	84.28\\
    + GRPO &7B  &  85.22&   \underline{82.89}&  84.29&  \underline{53.55}&  62.0 & 49.5 & 66.4 & 27.96 & 71.7 & 85.16  \\
    \rowcolor{gray!20}
    + Ours &7B  & \underline{86.09} & \textbf{86.84} &\underline{86.39} & \textbf{54.49} & \underline{63.8} &	\textbf{50.0}	& \textbf{67.4} &	\underline{30.92} &	72.0 &	84.44\\
    \bottomrule
    \end{tabular}}
\end{table*}

\begin{table}[tbp]
    \small
    \centering
    \label{tab:tts}
    \caption{Comparison of different test-time scaling strategies, where Truncate and Truncate-Thinking denote our proposed Truncate–then–Regenerate and Truncate–Thinking–then–Regenerate methods, respectively.}
    \scalebox{0.9}{
    \begin{tabular}{llcccccc}
    \toprule
    \multirow{2.5}{*}{\textbf{Sample}} &\multirow{2.5}{*}{\textbf{Method}} & \multicolumn{3}{c}{\textbf{V$^*$}} & \multirow{2.5}{*}{\textbf{BLINK}} \\
    \cmidrule(lr){3-5}
    & & \textbf{Attr} & \textbf{Pos} & \textbf{All} & \\
    \midrule
    \multirow{3}{*}{k=4} & Major voting & 91.30 &	76.32 &	85.34 &  48.24 \\
    & Truncate & 93.04 &	\textbf{77.63} &	\textbf{87.96} & \textbf{49.13}  \\
    & Truncate-Thinking &  \textbf{94.78} &	76.32 &	86.91 & 48.85\\
    
    \midrule
    \multirow{3}{*}{k=8} & Major voting & 92.17 &	76.32 &	85.86 & 48.41  \\
    & Truncate &  93.91 &	\textbf{78.95} &	\textbf{87.96} & \textbf{49.25} \\
    & Truncate-Thinking &  \textbf{94.78} &77.63 &  \textbf{87.96} & \textbf{49.25}\\ 
    
    \midrule
    \multirow{3}{*}{k=16} & Major voting & 92.17 &	76.32 &	85.86 & 48.41\\
     & Truncate & \textbf{94.78} & \textbf{81.57} & \textbf{89.53} & \textbf{49.45}  \\
    & Truncate-Thinking & \textbf{94.78} & 78.95 & 88.48 & 49.38\\

    \bottomrule
    \end{tabular}}
    
    \label{tab:tts}
\end{table}

\subsection{Main Results}
\paragraph{RL Training with PRM.}
As shown in Table~\ref{tab:main}, our method significantly and consistently outperforms the GRPO baseline across both 3B and 7B model scales. 
Specifically, for the 3B model, our approach achieves average improvements of approximately 4\% in the Visual Search category, 3\% in Math and Chart reasoning, and 1\% in Perception-intensive Reasoning relative to the GRPO baseline.
This result strongly demonstrates that our method provides richer and more fine-grained supervision. A deeper analysis of the Visual Search sub-tasks reveals that the most substantial gains originate from \textbf{$V^*_{pos}$} (Positional Perception), particularly at the 3B scale (e.g., improving from 86.95 to 90.43). This strongly suggests that our fine-grained process supervision has successfully guided the model to enhance its precise spatial localization capabilities. Concurrently, the improvements on benchmarks like BLINK and MMStar  also indicate that this enhanced perception leads to higher fidelity and fewer hallucinations.
A crucial finding is the model's strong generalization ability. As discussed in Section~\ref{sec:implementation_details}, although our PRM training and RL intervention were predominantly focused on Visual Search tasks, the model still exhibits consistent performance gains across all other domains, including general perception and math reasoning. We attribute this ``capability transfer'' to the fact that tasks in Math \& Chart (such as MathVision and ChartQA) are fundamentally reliant on precise, fine-grained perceptual abilities~(\eg, localizing data points on a chart, reading text). By strengthening the model's foundational perceptual accuracy, our method successfully generalizes this improvement to broader and more complex reasoning tasks.
Furthermore, our 7B model trained with our method also surpasses Pixel-Reasoner and achieves performance competitive with DeepEyes on Visual Search tasks. It is noteworthy that the latter two models both rely on external tool manipulation to assist in object grounding. This result indicates that enhancing the intrinsic perceptual abilities of multimodal base models is a highly promising research direction, capable of rivaling the performance of tool-augmented SOTA methods.

\paragraph{Test-time Scaling with PRM.}
As mentioned earlier, \textsc{Perceval} has the potential to assist in the test-time scaling of policy models with the Truncate or Feedback strategies. To validate their effectiveness, we compare them with the major voting strategy, a classic test-time scaling method, where the policy model generate responses for multiple times and selects the most common answer as the final response. We conducted the experiment on the 3B policy model and present the results in Table~\ref{tab:tts}. With different sampling times $k$, the PRM-based strategies consistently outperform major voting on V* and BLINK. The Truncate strategy, in particular, shows a more significant improvement compared to the Feedback strategy. We hypothesize that the model's training data does not contain sufficient reflective data, which results in poorer instruction-following quality when the reflective prompts are inserted in the Feedback strategy. In contrast, the Truncate strategy allows the model to regenerate the response based on its own generated context, aligning more closely with the model's original distribution, thus producing more stable and reliable outputs. Another observation is that the major voting strategy quickly converges on difficult tasks (\eg, the Pos subset of V*) and fails to show further improvement. This suggests that without external intervention, the model's inherent capabilities are insufficient to rectify its errors.


\subsection{Further Analysis}
\paragraph{Reward Hacking Test.}
\begin{figure}[tbp]
    \centering
    \includegraphics[width=\columnwidth]{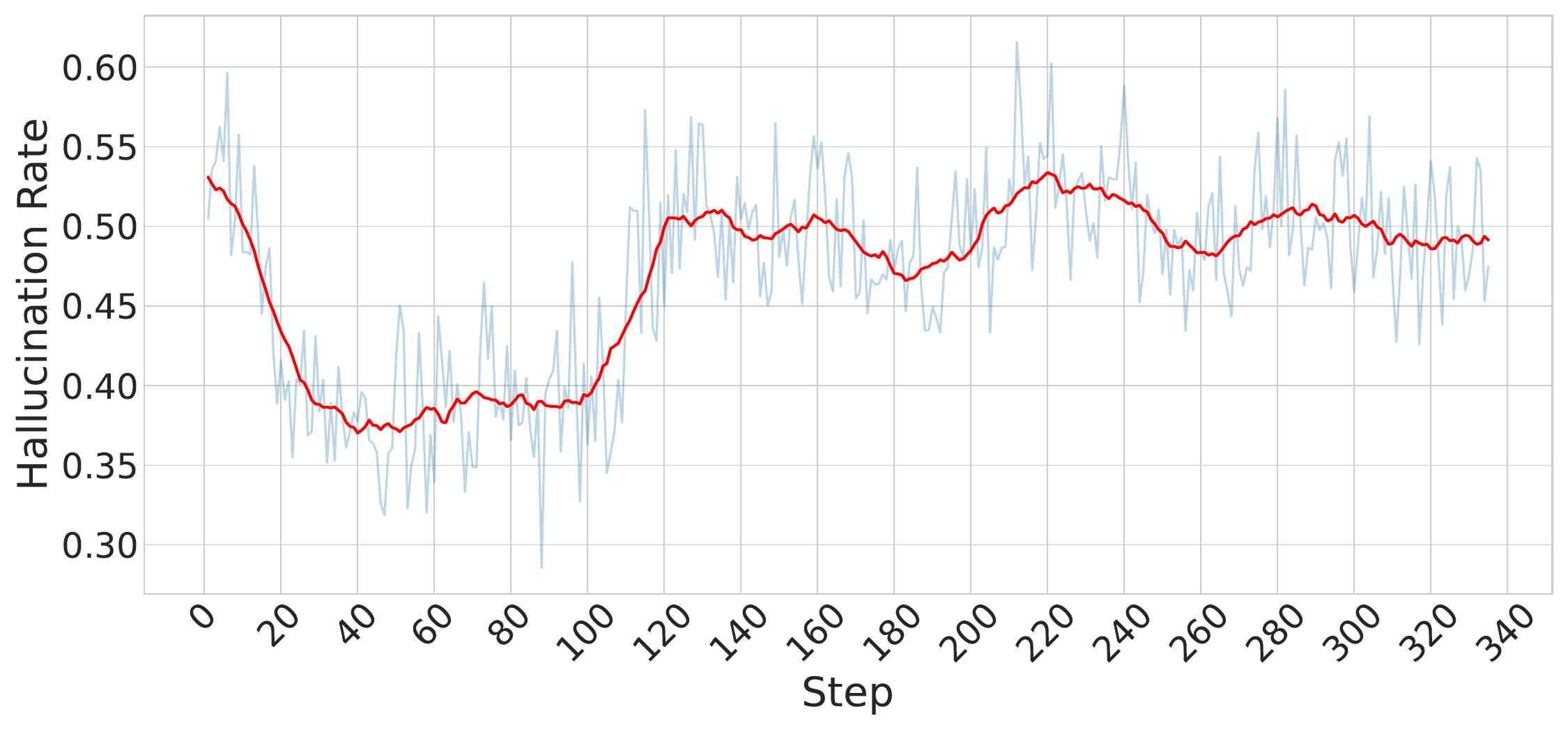}
    \caption{The proportion of responses identified by \textsc{Perceval} as containing hallucinations during training.}
    \label{fig:hallu_rate}
\end{figure}

A critical challenge in reinforcement learning with reward models (RMs) is \textit{reward hacking}, where the policy overfits the RM's scoring function. This issue is particularly pronounced with traditional RMs that output a single scalar reward for an entire response. Such a direct and holistic score, which is often influenced by the RM's own intrinsic biases, provides a simple signal for the policy to exploit, leading to score inflation without genuine quality improvement. Our proposed \textsc{Perceval} is designed to mitigate this specific vulnerability. Instead of providing a direct scalar reward, \textsc{Perceval} intervenes during the advantage calculation stage. Specifically, it reduces the advantage values of only those tokens within a response that are identified as contributing to a hallucination. This fine-grained, indirect guidance mechanism is inherently more difficult for the policy to overfit and simultaneously enhances the contrast between correct and incorrect tokens within the same sequence.
The effectiveness of this approach is demonstrated in Figure~\ref{fig:hallu_rate}, which plots the proportion of responses identified by \textsc{Perceval} as containing hallucinations during training. The curve initially shows a decline, indicating that the policy is successfully learning to reduce hallucinations. Crucially, the rate then stabilizes rather than continuing to drop. A continuously decreasing curve would suggest that the policy is learning to deceive the PRM---a clear sign of reward hacking. The observed stability therefore confirms that our proposed \textsc{Perceval} effectively guides the policy toward genuine improvement while avoiding significant reward hacking.


\begin{table}[tbp]
\centering
\small
\caption{Ablation study on the penalty strength hyperparameter $\alpha$.}
\label{tab:alpha_ablation}
\begin{tabular}{@{}ccccc@{}}
\toprule
\textbf{$\alpha$} & \textbf{V$^*$} & \textbf{RealWorldQA}  & \textbf{MathVision}  & \textbf{ChartQA}    \\
\midrule
0.0 & 80.10 & 62.17 & 23.36 & 83.32 \\ 
0.03 & 81.68  & 63.09 & 22.70 & 84.44 \\
0.1 & \textbf{83.25}  & \textbf{64.92} & \textbf{26.32} & \textbf{85.04} \\
0.3 & 78.53  & 61.78 & 22.04 & 84.56 \\
\bottomrule
\end{tabular}%
\end{table}

\paragraph{Hyperparameter Tuning.} Our proposed RL training with PRM framework introduces the hyperparameter $\alpha$ (Equation~\ref{eq:token_advantage}), which governs the penalty strength applied to tokens identified as hallucinatory. The selection of an optimal $\alpha$ is critical, as it requires balancing the suppression of hallucinations against the preservation of overall response quality. To quantitatively determine this optimal value, we conduct a series of experiments, varying $\alpha$ across $\{0.03, 0.1, 0.3\}$ and benchmarking against a standard GRPO baseline ($\alpha=0$). The results, summarized in Table~\ref{tab:alpha_ablation}, reveal a distinct non-monotonic trend.
A minimal value of $\alpha=0.03$ provides an insufficient corrective gradient. While offering a marginal improvement over the baseline, the penalty is too subtle to effectively steer the model away from ingrained hallucinatory patterns. Conversely, an excessively large $\alpha$ of 0.3 proves counterproductive. We attribute this to collateral ``penalization'': since the \textsc{Perceval} flags entire substrings, a high penalty indiscriminately punishes all tokens within that span, including syntactically necessary but factually benign words (e.g., articles, prepositions). This introduces significant training noise and degrades overall performance. The analysis reveals that $\alpha=0.1$ strikes the optimal balance. It is potent enough to achieve a substantial reduction in hallucinations
while avoiding the destabilizing effects of over-penalization. Therefore, we adopt $\alpha=0.1$ as the canonical value for all other experiments.

\begin{figure}
    \centering
    \includegraphics[width=1.0\linewidth]{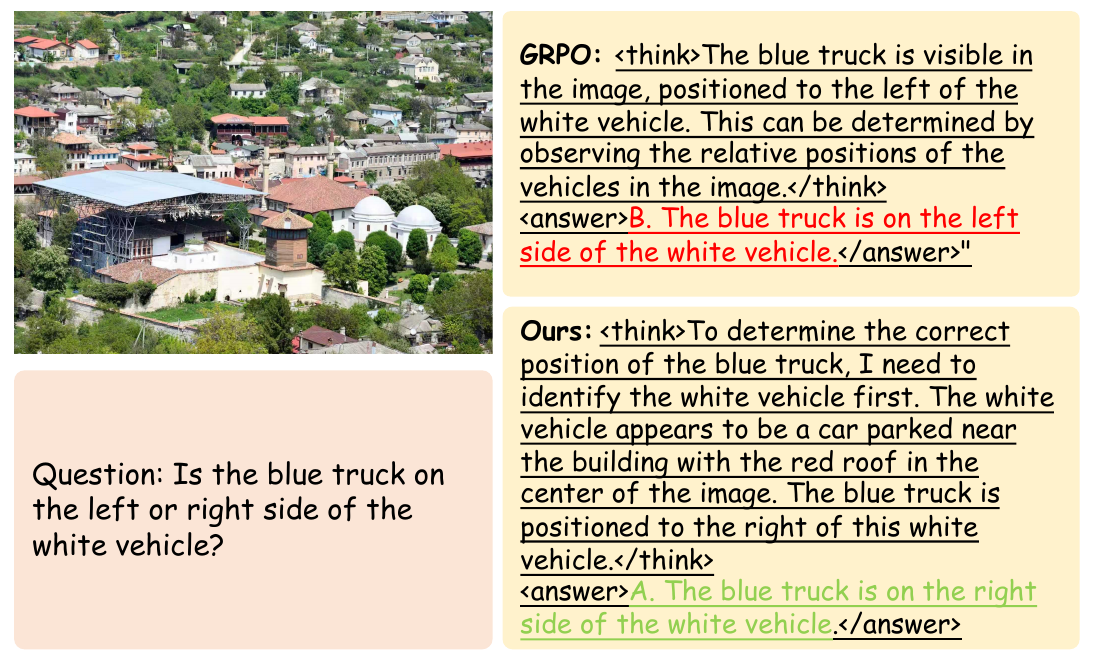}
    \caption{Case study of the visual reasoning process from models trained with GRPO and our method.}
    \label{fig:case}
\end{figure}

\paragraph{Qualitative Analysis.}
To clearly demonstrate the efficacy of our method, we present a qualitative analysis of model outputs in Figure~\ref{fig:case}. This case study compares the outputs from a model trained with direct GRPO against one trained with our method on an identical query. The task necessitates locating two minuscule objects~(\ie, a blue vehicle  and a vehicle car) to determine their spatial relationship.
The baseline model, trained with direct GRPO, bypasses the perceptual task and directly outputs a relative position (``left''). This is a classic example of hallucination, as the model provides an answer without seemingly grounding its response in the visual evidence.
In sharp contrast, our model exhibits a deliberate, step-by-step process. It first attempts to locate the white car, subsequently searches for the blue car, and then correctly deduces their relative positions. This case study demonstrates that our RL training process significantly enhances the model's perceptual capabilities, compelling its responses to be faithfully grounded in the visual content.

\section{Related Work}
\label{sec:related_work}

\paragraph{Vision-language Models}
The field of vision-language models (VLMs) has evolved from foundational representation alignment to complex multimodal reasoning. Early breakthroughs such as CLIP~\cite{radford2021learning} and ALIGN~\cite{jia2021scaling} demonstrate that contrastive pre-training on web-scale image-text pairs yields powerful, transferable representations, setting the stage for Large Vision Language Models (LVLMs) that bridge pre-trained visual encoders with LLMs~\cite{alayrac2022flamingo, li2023blip, liu2023visual}.
``Visual Instruction Tuning''~\cite{liu2023visual} emerges as a critical paradigm for unlocking multimodal instruction-following, rapidly scaled in open-source models like Qwen-VL~\cite{bai2023qwen} and InternVL~\cite{zhu2025internvl3}. By incorporating large-scale SFT and RL, advanced VLMs~\cite{yang2025qwen3, geminiteam2025geminifamilyhighlycapable} achieve strong performance on complex reasoning tasks. However, perceptual capabilities remain a critical bottleneck: models frequently exhibit hallucinations~\cite{li2023evaluating, li2025analyzing} or are unduly dominated by textual priors~\cite{zhang2025modalities, agrawal2025towards, liu2025more}, highlighting a persistent gap in reliable, fine-grained visual perception.

\vspace{-2pt}

\paragraph{Reinforcement Learning for VLMs}
The application of RL to VLMs has rapidly evolved toward capability incentivization for complex multimodal reasoning. This shift was catalyzed by breakthroughs in LLMs demonstrating that large-scale RL can elicit emergent ``slow-thinking'' behaviors~\cite{guo2025deepseek, jaech2024openai, team2025kimi}, inspiring a new wave of VLM research that optimizes the synergy between visual perception and logical deliberation~\cite{peng2025lmm, huang2025vision, shen2025vlm}. Beyond adapting LLM strategies, researchers explore reflection techniques tailored to the visual domain and ``thinking with images'' paradigms that leverage image manipulation tools to support reasoning. However, a critical limitation persists: methods based on RLVR predominantly rely on GRPO, which provides only coarse, outcome-level supervision and lacks the fine-grained signals necessary for improving complex, step-by-step reasoning.

\paragraph{Multimodal Reward Models}
Multimodal reward models~\cite{wang2025unifiedmultimodalchainofthoughtreward, zang2025internlmxcomposer25rewardsimpleeffectivemultimodal, zhang2025baserewardstrongbaselinemultimodal} play a pivotal role in Reinforcement Learning from Human Feedback (RLHF) by aligning model outputs with human preferences. With the recent proliferation of reinforcement learning in complex reasoning tasks, RMs are also increasingly employed to supplement methods like Reinforcement Learning with Verifiable Rewards (RLVR). This becomes particularly crucial in domains where verifiable ground truth is inaccessible, such as open-ended creative tasks~\cite{liu2025inferencetimescalinggeneralistreward}, which are environments where methods reliant on verifiable rewards consequently struggle.
The predominant approach for these RMs involves training them to directly output a single scalar score, which represents the overall quality of a given trajectory~\cite{zang2025internlmxcomposer25rewardsimpleeffectivemultimodal, wang2024math}. Recognizing the limitations of this direct scoring, more recent research efforts have sought to integrate ``slow thinking'' or deliberate reasoning paradigms into reward modeling~\cite{zhang2025r1rewardtrainingmultimodalreward, zhang2025structvrmaligningmultimodalreasoning}. These approaches enable the RM to generate a rationale or critique before assigning the final score, aiming for more meticulous and robust evaluations~\cite{fan2025sophiavlr1reinforcingmllmsreasoning}. However, a fundamental limitation persists: whether generated directly or after deliberation, the feedback from existing RMs ultimately collapses into a single scalar reward. This offers only sparse, outcome-level supervision for algorithms like GRPO. We propose a perception-centric reward model that provides a more fine-grained signal, which enabling token-level adjustments of advantages, thereby offering a more precise supervision.

\section{Conclusion}
In this work, we introduced \textsc{Perceval}, a perception-centric process reward model (PRM) that addresses the sparse reward issue in RLVR by enabling token-level error grounding. Unlike traditional outcome-level supervision, \textsc{Perceval} detects image–text misalignments within the model's reasoning process and provides grounded, step-aware feedback. We trained \textsc{Perceval} with perception-intensive data and integrate it into both the training and inference stages of VLMs. At the training stage, we leverage \textsc{Perceval} to apply token-level penalties to hallucinated spans, improving fine-grained credit assignment and surpassing the capabilities of sequence-level methods like GRPO. During inference, \textsc{Perceval} enables a Truncation–Regeneration loop that prunes erroneous responses and induces model reflection. Our experiments demonstrate that \textsc{Perceval} substantially improves visual grounding on perception-heavy benchmarks and facilitates better transfer to multi-step reasoning tasks. \ This method represents a significant advancement in fine-tuning the reasoning capabilities of VLMs, with the potential to generalize across domains and tasks.

\section{Acknowledge}

This work was partially supported by the National Natural Science Foundation of China No. 92470205 and Beijing Major Science and Technology Project under Contract No. Z251100008425002.
{
    \small
    \bibliographystyle{ieeenat_fullname}
    \bibliography{main}
}



\end{document}